\documentclass[10pt, journal, twocolumn]{IEEEtran}

\usepackage{cite}
\usepackage{graphicx}

\usepackage{amsmath}
\usepackage{booktabs}

\usepackage[colorlinks=true,
            linkcolor=blue,      
            citecolor=black,     
            urlcolor=blue]       
           {hyperref}

\usepackage{multirow}

\begin{document}

\title{Machine learning-enhanced non-amnestic Alzheimer's disease diagnosis from MRI and clinical features}

\author{Megan A. Witherow,$^1$ Michael L. Evans,$^1$ Ahmed Temtam,$^1$ Hamid R. Okhravi,$^2$ and Khan M. Iftekharuddin,$^1$ for the Alzheimer's Disease Neuroimaging Initiative$^*$

\thanks{1 Vision Lab, Department of Electrical and Computer Engineering, Old Dominion University, Norfolk, VA, USA 23529}
\thanks{2 Glennan Center for Geriatrics and Gerontology, Department of Medicine, Macon \& Joan Brock Virginia Health Sciences at Old Dominion University, Norfolk, VA, USA 23529}
\thanks{* Data used in preparation of this article were obtained from the Alzheimer's Disease Neuroimaging Initiative (ADNI) database (adni.loni.usc.edu). As such, the investigators within the ADNI contributed to the design and implementation of ADNI and/or provided data but did not participate in the analysis or writing of this report. A complete listing of ADNI investigators can be found at: http://adni.loni.usc.edu/wp-content/uploads/how\_to\_apply/ADNI\_Acknowledgement\_List.pdf 
}
\thanks{Correspondence to: Khan M. Iftekharuddin, \href{mailto:kiftekha@odu.edu}{\textcolor{blue}{kiftekha@odu.edu}}, 2131K Engineering Systems Building, Norfolk, VA, USA 23529}}

\maketitle

\begin{abstract}
Alzheimer’s disease (AD), defined as an abnormal buildup of amyloid plaques and tau tangles in the brain can be diagnosed with high accuracy based on protein biomarkers via PET or CSF analysis. However, due to the invasive nature of biomarker collection, most AD diagnoses are made in memory clinics using cognitive tests and evaluation of hippocampal atrophy based on MRI. While clinical assessment and hippocampal volume show high diagnostic accuracy for amnestic or typical AD (tAD), a substantial subgroup of AD patients with atypical presentation (atAD) are routinely misdiagnosed. To improve diagnosis of atAD patients, we propose a machine learning approach to distinguish between atAD and non-AD cognitive impairment using clinical testing battery and MRI data collected as standard-of-care. We develop and evaluate our approach using 1410 subjects across four groups (273 tAD, 184 atAD, 235 non-AD, and 685 cognitively normal) collected from one private data set and two public data sets from the National Alzheimer’s Coordinating Center (NACC) and the Alzheimer’s Disease Neuroimaging Initiative (ADNI). We perform multiple atAD vs. non-AD classification experiments using clinical features and hippocampal volume as well as a comprehensive set of MRI features from across the brain. The best performance is achieved by incorporating additional important MRI features, which outperforms using hippocampal volume alone.  Furthermore, we use the Boruta statistical approach to identify and visualize significant brain regions distinguishing between diagnostic groups. Our ML approach improves the percentage of correctly diagnosed atAD cases (the recall) from 52\% to 69\% for NACC and from 34\% to 77\% for ADNI, while achieving high precision. The proposed approach has important implications for improving diagnostic accuracy for non-amnestic atAD in clinical settings using only clinical testing battery and MRI. 
\end{abstract}


\IEEEpeerreviewmaketitle

\section{Introduction}

\IEEEPARstart{A}{lzheimer’s} disease (AD) is a progressive, neurodegenerative condition associated with abnormal buildup of amyloid plaques and tau tangles in the brain, leading to brain atrophy and cognitive decline. While typical AD (tAD) presents with episodic memory loss and hippocampal atrophy, a substantial subgroup (at least 25\%) of all AD cases present without memory deficits, especially in the early stages of the disease \cite{RN28,RN27,RN25}. Rather, these non-amnestic, atypical AD (atAD) patients present with deficits in other cognitive domains, such as language, visuospatial, behavioral, or motor functioning and are associated with relative sparing of the hippocampus \cite{RN28,RN27,RN25}. Since AD is usually diagnosed based on MRI-derived hippocampal volume and a battery of cognitive tests that assess memory, many of these patients experience delays in diagnosis and may be misdiagnosed with other dementias or neurological conditions causing non-AD cognitive impairment (non-AD). Diagnostic confirmation requires more comprehensive testing such as lumbar puncture to obtain a CSF sample for protein analysis or PET scan with radiotracers that bind to amyloid and tau aggregates in the brain \cite{RN27}. While effective, these procedures are more expensive, invasive, and carry higher risk (e.g., radiation exposure during nuclear imaging) \cite{RN27}. Therefore, there is a need for methods that increase atAD diagnostic accuracy using clinical measurements collected as standard-of-care. \par

While tAD is primarily associated with hippocampal atrophy, atAD has multiple variants, each associated with different patterns of neurodegeneration. Behavioral or frontal variant of atAD is associated with frontal and anterior temporal regions, and may present with symptoms like those of the behavioral variant of frontotemporal dementia, leading to misdiagnosis \cite{RN25}. The language variant, also known as logopenic progressive aphasia due to AD, is associated with greater rates of atrophy in the left compared to the right hemisphere of the brain, beginning with the temporoparietal junction, posterior cingulate, and precuneus and progressing into temporal, frontal, and caudate regions \cite{RN28, RN27}. Patients with the visuospatial variant of atAD may meet diagnostic criteria for posterior cortical atrophy, a neurodegenerative condition associated with difficulties in vision, spatial processing, and perception \cite{RN25}. While AD is the primary cause of posterior cortical atrophy, other causes may include dementia with Lewy bodies, prion disease, and corticobasal degeneration \cite{RN25}. MRIs of patients with the visuospatial variant show atrophy in posterior parietal and occipital regions, with initial sparing of the hippocampus and other areas within the medial temporal lobe \cite{RN25}. The motor variant of atAD, associated with temporoparietal, posterior cingulate, basal ganglia, thalamus, insula, and occipital lobe atrophy, may manifest as corticobasal syndrome, with clinical features such as Parkinsonism, dystonia, limb apraxia, and others \cite{RN25}. The diffuse selection of brain regions involved in atAD variants suggests that MRI features from brain regions in addition to the hippocampus may provide additional and complementary information to inform diagnosis of atAD. \par

Machine learning (ML) and pattern-recognition techniques have found wide application in neuroimaging for discovery of previously unknown patterns in imaging data. In recent years, several studies have applied ML for early detection and classification of AD.  Imaging data have been used with ML to detect early onset AD \cite{RN22}. For example, in Yue et al.,\cite{RN23} brain feature maps are input into a deep ML model to learn hidden features, achieving F1 scores of over 98\% during a binary classification task on the Alzheimer’s Disease Neuroimaging Initiative (ADNI)-2 dataset. ML models trained on medical imaging data can be used to diagnose AD during early stages of cognitive impairment and help identify which brain regions play a significant role in disease progression. In a study by Fujita et al.,\cite{RN26} ML models are trained for AD vs. non-AD classification using cognitive test scores, patient educational level, and z-scores quantifying the atrophy of the medial temporal area of the brain, including the hippocampus, amygdala, and entorhinal cortex. \par

However, there has been limited work to improve diagnosis of difficult to diagnose atAD cases. Our previous work \cite{RN21} demonstrates that ML can classify tAD vs. non-AD, atAD vs. non-AD, and combined tAD+atAD vs. non-AD using clinical features and hippocampal volumes collected from 173 patients from a single tertiary memory center site. In this manuscript, we build upon these preliminary results to develop and evaluate a more comprehensive ML-based diagnostic pipeline using a larger cohort of subjects from public sources, including ADNI and the National Alzheimer’s Coordinating Committee (NACC) Uniform Data Set, as well as a data set of memory clinic patients.  Specific contributions of this work are as follows:

\begin{itemize}
    \item Identify non-amnestic atAD cases with MRI scans within different public data sets (NACC and ADNI) based on logical memory scores and build a dataset. From the MRI scans, we extract volumes, cortical thicknesses, and surface areas from 34 regions of interest per hemisphere of the brain.
    \item Develop Random Forest models to classify between atAD and non-AD cases using features currently used in clinical practice including cognitive test scores and hippocampal volume.
    \item Study the contributions of clinical and MRI features from across the whole brain using an ablation study.
    \item Identify and visualize significant brain regions for distinguishing between diagnostic groups to further characterize tAD, atAD, and non-AD groups and highlight the most important features for atAD diagnosis. 
\end{itemize}

\section{Materials and methods}

\subsection{Data}
Data used in the preparation of this study are obtained from three data sets. A private set with de-identified data is obtained from the Glennan Center for Geriatrics and Gerontology at Brock Virginia Health Science (VHS), Old Dominion University. This study has been reviewed and approved by the VHS IRB under IRB \#18-08-WC-0215. Data are also obtained from two public data sets, the NACC Uniform Data Set Version 3 \cite{RN36,RN9} and ADNI database \cite{RN37,RN10}. These three data sets are described in further detail below.

\subsubsection{VHS data set}
The VHS data set consists of de-identified data from 177 patients aged 65-90 who are evaluated at the VHS Glennan Center Memory Center. All patients had baseline visits from January 1, 2015 through September 2024. Diagnostic work up for each participant includes a 120-minute standardized baseline assessment by interviewing and examining the patients, administration of a 90-minute battery of neuropsychological tests that assess all major cognitive domains, and interview of a proxy close to the patient. Clinical diagnosis is made by a cognitive disorder specialist. Patients also undergo MRI imaging of the brain and volumetric analysis of the hippocampus by commercially available fully-automated software NeuroQuant®. Diagnosis of probable AD at the stages 3-6 (MCI or dementia with differing severity) is made based on clinical, MRI, PET, and CSF data, based on the National Institute on Aging and Alzheimer’s Association core clinical criteria \cite{RN35,RN34}.

\subsubsection{NACC data set}
The NACC Uniform Data Set \cite{RN9} includes participants from the National Institute on Aging’s Alzheimer’s Disease Research Centers Program since 2005. The data includes 18 data collection forms completed by clinicians on family history, risk factors, and neurophysiological battery. The NACC data set includes subjects with AD, other diagnoses causing cognitive impairment, and healthy subjects. MRI scans and biomarker testing results are available for a subset of the patients. For more information, please visit NACC’s website \cite{RN36}.

\subsubsection{ADNI data set}
ADNI \cite{RN10}, launched in 2003, is a public-private partnership with the original goal of testing whether serial MRI, PET, other biological markers, and clinical and neuropsychological assessment can be combined to measure the progression of mild cognitive impairment (MCI) and early AD. Current goals include validating biomarkers for clinical trials, improving the generalizability of ADNI data by increasing diversity in the participant cohort, and providing data concerning the diagnosis and progression of AD to the scientific community. For up-to-date information, see adni.loni.usc.edu.

\begin{table}
    \centering\small
   \caption{\small Number of samples in each data set by group.
 \label{tab:tab1}}
    \begin{tabular}{c|c|c|c|c|c}
    \toprule
    {\bf Data Set} & {\bf tAD} &  {\bf atAD} & {\bf non-AD} & {\bf CN} & {\bf Total}\\
    \midrule
    NACC (Public) & 92 & 29 & 59 & 264 & 444 \\
    ADNI (Public) & 144 & 87 & 137 & 421 & 789 \\
    \textit{Public Total} & \textit{203} & \textit{116} & \textit{196} & \textit{685} & \textit{1233} \\
    VHS (Private) & 70 & 68 & 69 & 0 & 177 \\
    \textit{\textbf{Combined Total}} & \textit{\textbf{273}} & \textit{\textbf{184}} & \textit{\textbf{235}} & \textit{\textbf{685}} & \textit{\textbf{1410}} \\
    \bottomrule
    \end{tabular}
\end{table}

\begin{table*}
\centering
\small
\caption{Initial clinician diagnosis for biomarker-confirmed AD cases by data set (NACC or ADNI) and subtype (tAD or atAD).}
\label{tab:tab2}
\begin{tabular*}{\textwidth}{@{\extracolsep{\fill}} l l cc c c}
\toprule
\textbf{Data Set} & \textbf{Diagnostic Subgroup} & \multicolumn{2}{c}{\textbf{Initial Clinician Diagnosis}} & \textbf{Total \# AD} & \textbf{\% Correctly Diagnosed} \\
\cmidrule(lr){3-4}
 &  & \textbf{AD} & \textbf{non-AD} & \textbf{(Confirmed via biomarker testing)} & \textbf{(Recall $\times$ 100)} \\
\midrule
NACC & tAD & 84 & 8  & 92 & 91.30\% \\
     & atAD & 15 & 14 & 29 & 51.72\% \\
ADNI & tAD & 134 & 10 & 144 & 93.06\% \\
     & atAD & 30 & 57 & 87 & 34.48\% \\
\bottomrule
\end{tabular*}
\end{table*}

\subsection{Data selection criteria}
We organize subjects into four groups: tAD, atAD, non-AD cognitive impairment (non-AD), and cognitively normal (CN). Inclusion/exclusion criteria for each group of patients are described in the sections below. The distribution of patients in each group for each data set is shown in Table~\ref{tab:tab1}. \par

The NACC and ADNI data sets report the initial clinician diagnosis prior to biomarker testing. Diagnosis has been confirmed for all patients based on protein biomarkers from PET, CSF, or autopsy. Table~\ref{tab:tab2} reports the percentage of patients with correct initial clinician diagnosis (as reported in the NACC or ADNI databases) at baseline for the tAD and atAD groups. This percentage corresponds to the recall:

\begin{equation}
\mathrm{Recall} = \frac{TP}{TP + FN}
\label{eq:recall}
\end{equation}

where \textit{TP} is the number of ‘true positives’, i.e., patients with biomarker evidence of AD pathology that received an initial clinician diagnosis of AD, and \textit{FN} is the number of ‘false negatives’, i.e., patients with biomarker evidence of AD pathology that received an initial clinician diagnosis of nonAD. \par

\subsubsection{tAD and atAD}
Subjects in the tAD and atAD groups must have no reported traumatic brain injury and must have evidence of AD pathology based on PET scan, CSF analysis, or autopsy neuropathology results \cite{RN32,RN31,RN30,RN33,RN20,RN19,RN18,RN17,RN16,RN14,RN13,RN12,RN11}. For the VHS data set, subjects undergo either PET scan or CSF analysis. PET scans are visually read by experts, which has been shown to be comparable to quantitative readings \cite{RN32,RN31,RN30}. CSF analysis is conducted using the Mayo clinic ADEVL test and associated cutoffs \cite{RN33}. \par

For the ADNI and NACC data sets, subjects that have undergone amyloid and tau PET scans for AD diagnosis must be amyloid beta positive based on SUVR $\geq$ 1.48 (centiloid 22) and tau positive based on temporal lobe SUVR $\geq$ 1.29 \cite{RN20,RN19,RN18,RN17,RN16}. Subjects that have undergone lumbar puncture and CSF analysis for AD diagnosis must have protein concentrations of amyloid beta 42 below 700 pg/ml, total tau above 400 pg/ml, and phosphorated tau above 60 pg/ml \cite{RN14}. Subjects that have neuropathology results based on autopsy must have a Consortium to Establish a Registry for Alzheimer's Disease (CERAD) score of 2 or 3 (moderate or frequent neural plaques) and Braak stage for neurofibrillary degeneration $\geq$ 3 \cite{RN13}. \par

For the VHS data set, patients have been divided into tAD and atAD groups based on their amnestic or non-amnestic presentation, respectively, during clinical assessment. For NACC and ADNI data sets, subjects with AD are further divided into tAD (amnestic) and atAD (non-amnestic) subgroups based on their logical memory (LM) scores from Wechsler Memory Scale-Revised test \cite{RN12}. Non-amnestic cases have a higher LM score than education thresholds \cite{RN11} defined for each group as follows. \par

\subsubsection{tAD}
Subjects with 16 or more years of education and LM Score $<$ 9, 8-15 years of education and LM Score $<$ 5, or 0-7 years of education and LM Score $<$ 3 are included in the tAD group. \par

\subsubsection{atAD}
Subjects with 16 or more years of education and LM Score $\geq$ 9, 8-15 years of education and LM Score $\geq$ 5, or 0-7 years of education and LM Score $\geq$ 3 are included in the atAD group.

\subsubsection{non-AD}
Subjects in the non-AD group must have no reported traumatic brain injury and have received a diagnosis of cognitive impairment. Excluding traumatic brain injury, there are no limitations on the presumptive etiology of the cognitive impairment, with examples including vascular dementia, frontotemporal dementia, corticobasal degeneration, dementia with Lewy bodies, Parkinson’s disease dementia, prion disease, alcohol-related dementia, HIV, and others. Patients must have tested negative for amyloid beta and/or tau pathology based on PET scan, CSF analysis, or autopsy. \par

\subsubsection{CN}
Subjects in this group must have no reported traumatic brain injury and no diagnosis of cognitive impairment. Patients must have tested negative for amyloid beta and/or tau pathology based on PET scan, CSF analysis, or autopsy. \par

\subsection{Clinical features and battery of cognitive tests}
Clinical features used in our analysis include the subject age, years of education, and whether the subject has a family history of dementia. In addition, we include the total score and sub-scores of five different cognitive tests commonly used in the diagnosis of AD. These cognitive tests are described in the sections below. \par

\subsubsection{Montreal Cognitive Assessment (MoCA)}
The Montreal Cognitive Assessment (MoCA) is a rapid, 30-point screening instrument used to detect early dementia and mild cognitive impairment. The assessment includes both verbal and written components and evaluates participants on the following sub-scores: Visuospatial Executive, Naming, Attention, Language, Abstraction, Memory, and Orientation. The MoCA has demonstrated strong positive and negative predictive values for both Alzheimer’s disease and MCI. \par

\subsubsection{Mini-Mental State Examination (MMSE)}
The Mini-Mental State Examination (MMSE) is a short screening tool to assess an overall measure of cognitive impairment. It is a set of 11 questions, scored out of 30, and evaluates orientation to time, orientation to place, and drawing. \par

\subsubsection{Trail Making Test (TMT) A \& B}
The Trail Making Test (TMT) is a timed, two-part assessment designed to evaluate rote memory and executive functioning. Test A requires the patient to connect a sequence of numbered circles in order as quickly as possible using a pencil. In test B, the patient is required to connect numerical and alphabetical circles together in order, alternating between the two. A score is considered deficient if it takes 78 seconds or more to complete test A, or 273 seconds or more for test B. The examiner will correct mistakes made by the patient, which are included in the total time score. \par

\subsubsection{Animal fluency}
The animal fluency test \cite{RN4} is a simple, timed neuropsychological measurement test in which participants are given 60 seconds to name as many animals as possible. Patients unable to name 15 or more animals within the given time frame may be at risk for having early stages of dementia or amnestic mild cognitive impairment (aMCI) \cite{RN5}. Individuals with aMCI have been shown to perform significantly worse on the animal fluency test than cognitively normal patients \cite{RN5}. \par

\subsubsection{Boston Naming Test (BNT)}
The Boston Naming Test (BNT) \cite{RN3} is a brief quantitative measurement of confrontation naming ability. The examinee is asked to name 60 drawings of objects that decrease in familiarity as the test progresses. Conditions such as cerebrovascular accidents and subcortical diseases can contribute to poor performance on the BNT. The BNT is included in AD testing batteries as it can quickly identify aphasia resulting from dementia. \par

\subsection{MRI feature extraction}
Grey matter volume, cortical surface area, and cortical thickness of individually segmented brain structures are obtained using a FreeSurfer (FS) v.7.4 data processing pipeline. FS is an open-source segmentation tool that we use to label each voxel as one of 34 regions of interest per hemisphere according to the Desikan-Killiany atlas. Left and right hemisphere values were extracted independently and summed to obtain selected total brain volumes, including estimated total intracranial volume (eTIV). T1-weighted magnetic resonance images from the sagittal acquisition plane are used for the cortical reconstruction and subcortical segmentation processes. Series with the latest scan date are selected for patients with multiple scans fitting our criteria. \par

The Digital Imaging and Communications in Medicine (DICOM) files for each patient are converted into a single Neuroimaging Informatics Technology Initiative (NifTI) file format, and then into a mgz format through the mri\_convert FS command to store high-resolution structural data. Next, the recon-all automatic cortical reconstruction procedure is performed in parallel on each hemisphere on a high-performance computing cluster to reconstruct the three-dimensional volume into a two-dimensional cortical surface. Lastly, the numerical data is extracted following the segmentation process and rounded to 3 decimal places for the following regions: cerebrum total cranial volume, total grey matter volume, total hippocampus volume, white matter hyperintensity volume, and total hippocampus volume. \par

\subsection{MRI feature preprocessing}
For each MRI feature (volume, cortical thickness, or surface area), we linearly regress out the effects of age, sex, and total intracranial volume (TIV). Using healthy CN subject data, we fit generalized linear models (GLMs) to predict each MRI feature based on age, sex, and TIV. The GLM equation is as follows:

\begin{equation}
\mathrm{MRI\_feature} =
\alpha
+ \text{age} \cdot \beta_{\text{age}}
+ \text{sex} \cdot \beta_{\text{sex}}
+ \text{TIV} \cdot \beta_{\text{TIV}}
+ \epsilon,
\end{equation}

\noindent
where $\alpha$ is the intercept; $\beta_{\text{age}}$, $\beta_{\text{sex}}$, and $\beta_{\text{TIV}}$ are the coefficients for age, sex, and TIV, respectively; and $\epsilon$ represents the errors. We use the fitted GLMs to predict the expected value of each MRI feature for each subject as follows: 

\begin{equation}
{\mathrm{MRI\_feature}}_{\text{predicted}}
=
\hat{\alpha}
+ \text{age} \cdot \hat{\beta}_{\text{age}}
+ \text{sex} \cdot \hat{\beta}_{\text{sex}}
+ \text{TIV} \cdot \hat{\beta}_{\text{TIV}},
\end{equation}

\noindent
where $\hat{\alpha}$ is the estimated intercept and $\hat{\beta}_{\text{age}}$, $\hat{\beta}_{\text{sex}}$, and $\hat{\beta}_{\text{TIV}}$ are the estimated coefficients for age, sex, and TIV, respectively. Then, we compute the z-score as the difference of the observed and predicted feature values divided by the standard deviation (SD) of the CN residuals:

\begin{equation}
\text{z-score} =
\frac{\mathrm{MRI\_feature}_{\text{predicted}} - \mathrm{MRI\_feature}_{\text{observed}}}
     {\text{SD of CN residuals}}.
\end{equation}

We implement our GLMs in Python using the statsmodels library \cite{RN39}. \par

\subsection{Machine learning for diagnostic prediction}
Our goal is to improve atAD diagnostic prediction, i.e., atAD vs. non-AD classification, using ML. Our features include cognitive test scores and MRI-derived volumes, cortical thicknesses, and surface areas of brain regions in the Desikan-Killiany atlas. We select the Random Forest algorithm as it provides multiple advantages towards building our ML classification approach. First, Random Forest is robust to multicollinearity and suitable for handling the expected feature dependence among cognitive tests measuring the same or similar constructs and among related brain regions. Second, Random Forest does not assume any particular underlying distribution of the data. This flexibility in handling various data distributions is appropriate for our atAD group, which may be composed of multiple different variants, and non-AD group, which may be composed of multiple different diagnoses. We use the Scikit-learn \cite{RN38} implementation of the Random Forest algorithm. During hyperparameter tuning, a grid search is performed to select the maximum tree depth (1, 2, 3, 4, or 5 levels) and number of trees in the ensemble (two, five, ten, 100, or 1000 trees) to optimize the validation performance. \par

We perform two types of ML experiments. First, we assess how well ML performs in atAD vs. non-AD classification using only features currently used in clinical practice (family history, cognitive test scores, and hippocampal volume). We perform classification experiments using each data set individually (VHS, NACC, ADNI), the combination of the two public data sets (NACC and ADNI), and all three data sets combined. \par

Second, to study the utility of different types of features (e.g., clinical test scores, hippocampal volume, and comprehensive selection of MRI features), we perform an ablation study. Since we do not have access to the full set of MRI features for the VHS data set, we perform the ablation study using the two public data sets (NACC and ADNI). \par

For all experiments, we perform a 5x2 nested cross validation. The outer 5-folds are split into a training set (4 folds) and a test set (1 fold). The process is repeated five times such that each fold serves as the test set once. In the inner 2-fold cross validation loop, each training set is split into two inner folds (training and validation) that are used to tune the model hyperparameters. The best performing hyperparameters are used to train the model on the combined training set and the trained model is evaluated on the held-out test set. This 5x2 nested cross validation ensures that all samples are used in evaluation exactly once while preventing data leakage that may result in inflated assessments of the model performance. We report precision, recall, and F1 score (the harmonic mean of precision and recall) in atAD vs. non-AD classification for all experiments. \par

\subsection{Identification of significant brain regions}
To characterize our atAD subjects in comparison to tAD, non-AD, and CN groups, we study the importance of our MRI features from different brain regions in distinguishing among these groups using the Boruta statistical method \cite{RN2}. Boruta is a non-parametric approach based on Random Forest for selecting all relevant features given a specified Type I error rate.  The Boruta approach does not require any assumptions of normality or independence and works well even given correlated features, as it was originally developed for genetics research \cite{RN2}. Using Boruta, we identify all significant brain regions distinguishing between atAD vs. CN, tAD vs. CN, and non-AD vs. CN to characterize each group in comparison to the healthy population. Then, we identify significant regions for distinguishing between atAD and non-AD. To perform our analysis, we use the BorutaPy \cite{RN40} implementation of the Boruta algorithm with the following settings: \textit{n\_estimators= ‘auto’, alpha = 0.05, perc = 100, max\_iter = 1000, two\_step = False}. We limit the maximum tree depth to five to mitigate overfitting and weight each class based on its frequency in the training set to address class imbalance. \par 

\subsection{Visualization of significant brain regions}
An anatomically average T1-weighted brain image from the ICBM 152 template is used as the background for mapping our regions of interest (ROIs) defined by the Desikan-Killiany volumetric atlas. To visualize specific regions, we construct an ROI mask by initializing each voxel to 0 and selectively assigning a value of 1 or 2 to color the voxels corresponding to each ROI. Brain regions found to be statistically significant ($\alpha$ = 0.05) in distinguishing between the two patient groups are used as the ROIs. The group mean z-score of each feature determines the color of its corresponding voxels: blue indicates a lower feature score relative to the comparison group, while red indicates a higher score. The resulting mask is overlaid on the background image to emphasize these regions using the nilearn Python library. \par

\section{Results}

\subsection{Subject data distributions}
\begin{figure}[h]
\centering
\includegraphics[width=\columnwidth]{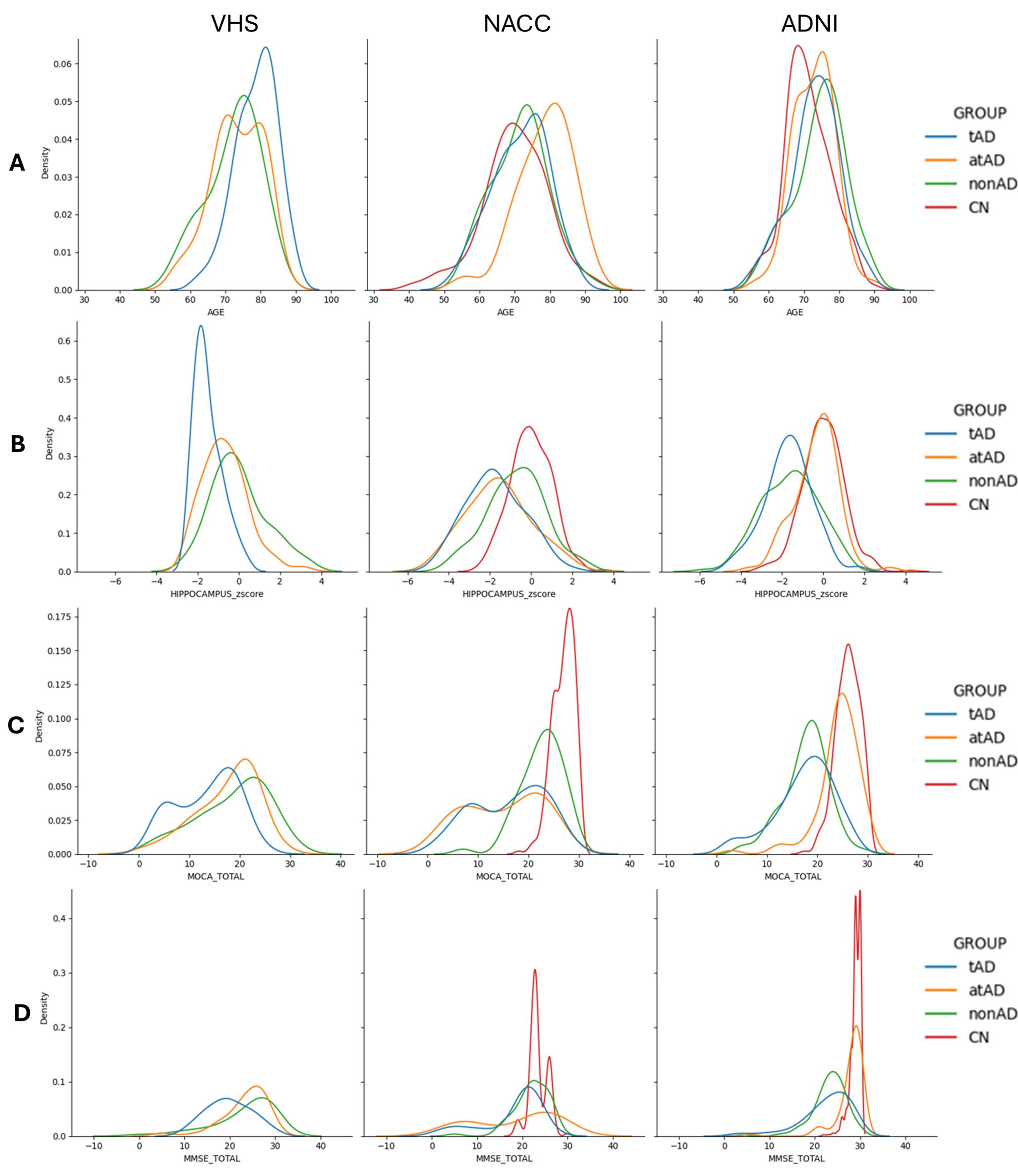}
\caption{Distribution of \textbf{(A)} subject ages, \textbf{(B)} z-score of hippocampal volume, \textbf{(C)} MoCA total scores, and \textbf{(D)} MMSE total scores for tAD, atAD, non-AD, and CN groups within the VHS (first column), NACC (second column), and ADNI (third column) data sets. Note that the VHS data set does not have a CN group.}
\label{fig:fig1}
\end{figure}

Fig.~\ref{fig:fig1}A-D plot the distributions of subject ages, normalized hippocampal volumes (z-scores), MoCA total scores, and MMSE total scores for the tAD, atAD, nonAD, and CN groups. Note that there is no CN group for the VHS data set. Across all data sets, the mean normalized hippocampal volume (Fig.~\ref{fig:fig1}B) for atAD is larger than tAD which is consistent with a non-amnestic (hippocampus-sparing) presentation. For VHS and ADNI data sets, the atAD group scores higher than tAD on average on the MoCA (Fig.~\ref{fig:fig1}C) and MMSE (Fig.~\ref{fig:fig1}D), as expected with a non-amnestic presentation. For NACC, atAD scores higher than tAD on average for MMSE and the groups are more closely aligned on MoCA score distributions. \par

\begin{table*}[t]
\centering
\small
\caption{Five-fold cross-validation classification results for atAD vs. non-AD considering different feature sets.}
\label{tab:tab4}
\begin{tabular*}{\textwidth}{@{\extracolsep{\fill}} l ccc ccc }
\toprule
\textbf{Data Set} & \multicolumn{3}{c}{\textbf{NACC}} & \multicolumn{3}{c}{\textbf{ADNI}} \\
\cmidrule(lr){2-4} \cmidrule(lr){5-7}
\textbf{Features} & Precision & Recall & F1 Score & Precision & Recall & F1 Score \\
\midrule
Hippocampal volume only       & 0.33 & 0.21 & 0.26 & 0.60 & 0.70 & 0.65 \\
Clinical only                 & 0.73 & 0.66 & 0.69 & 0.87 & 0.77 & 0.82 \\
Hippocampal volume + clinical & 0.68 & 0.59 & 0.63 & 0.86 & 0.77 & 0.81 \\
MRI only                      & 0.71 & 0.69 & 0.70 & 0.73 & 0.64 & 0.68 \\
\textbf{MRI + clinical}       & \textbf{0.77} & \textbf{0.69} & \textbf{0.73} & \textbf{0.89} & \textbf{0.77} & \textbf{0.83} \\
\bottomrule
\end{tabular*}
\end{table*}

\subsection{Machine learning experiments for atAD diagnosis}
\begin{figure}[h]
\centering
\includegraphics[width=\columnwidth]{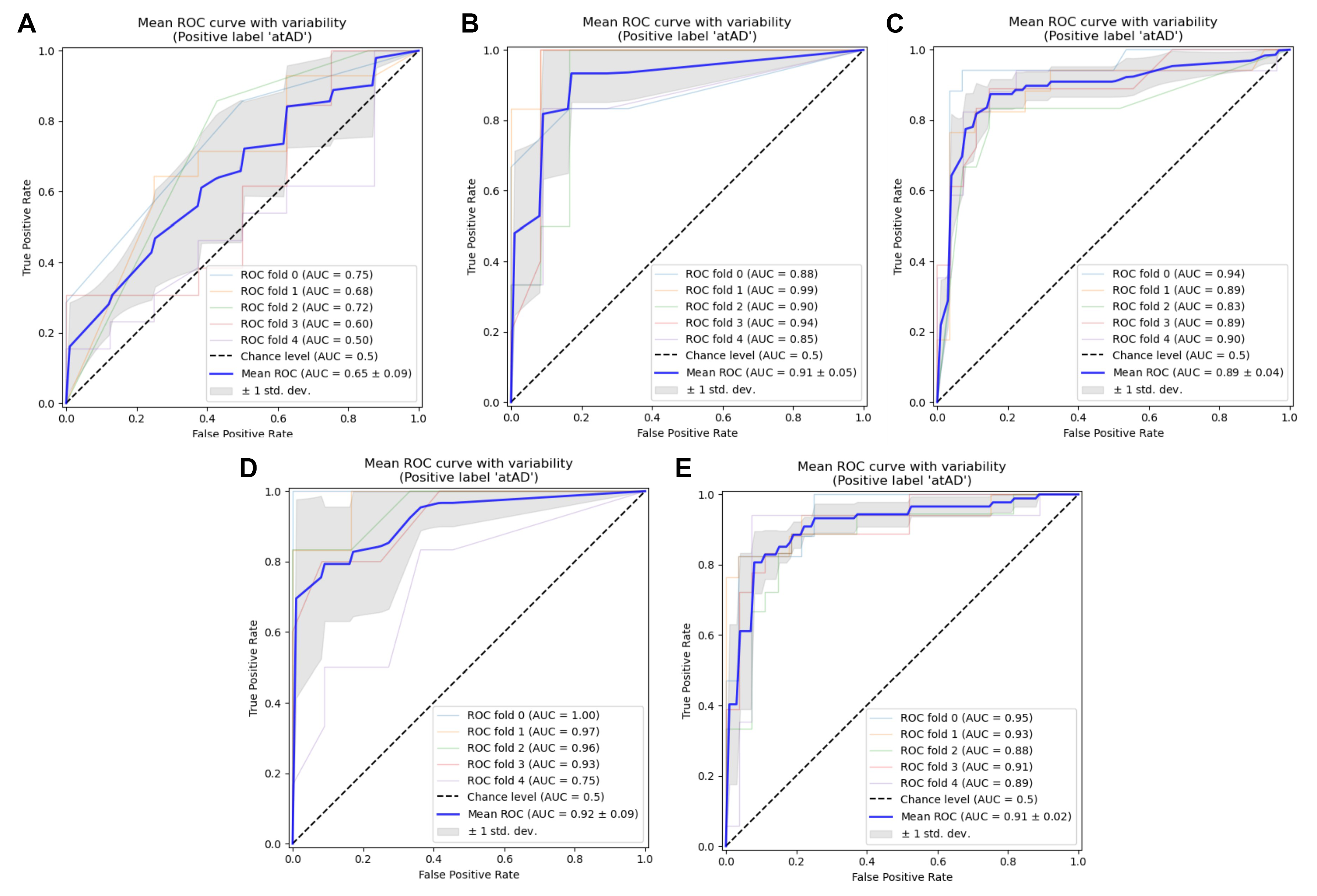}
\caption{ROC curves for 5-fold cross validation atAD vs. non-AD classification: (first row) clinical scores + hippocampal volume for \textbf{(A)} VHS data set, \textbf{(B)} NACC data set, and \textbf{(C)} ADNI data set; (second row) clinical scores + MRI features for \textbf{(D)} NACC data set and \textbf{(E)} ADNI data set.}
\label{fig:fig2}
\end{figure}

Table~\ref{tab:tab3} reports 5-fold cross-validation atAD vs. non-AD classification results using patient demographics, cognitive test scores, and hippocampal volume as features. The highest recall is achieved for the VHS data set (0.82), followed by ADNI (0.77), and NACC (0.59). Our results show substantial improvement over the baseline recall of 0.34 for ADNI. We also see some improvement of the baseline recall of 0.52 for NACC. Fig.~\ref{fig:fig2}A-C plots the receiver operating characteristic (ROC) curves and reports the area under the curve (AUC) for each of the five folds of the (A) VHS, (B) NACC, and (C) ADNI data sets. \par

Table~\ref{tab:tab4} reports the feature ablation study results to compare the contributions of clinical features, hippocampal volume, and MRI features. For both NACC and ADNI, the best performing feature set across all metrics is the combination of MRI and clinical features, achieving a recall of 0.69 for NACC and 0.77 for ADNI. The poorest performing feature set is hippocampal volume only. For both data sets, the results show that even clinical features alone are discriminative for distinguishing atAD from non-AD. These clinical features may then be further complimented by MRI. Fig.~\ref{fig:fig2}D and E shows ROC curves and AUCs for the five folds of the (D) NACC and (E) ADNI data sets. \par

\begin{table}
    \centering\small
   \caption{\small Five-fold cross-validation classification results for atAD vs. non-AD based on features currently used in practice (clinical scores + hippocampal volume).
 \label{tab:tab3}}
    \begin{tabular}{c|c|c|c}
    \toprule
    {\bf Data Set} & {\bf Precision} &  {\bf Recall} & {\bf F1 Score}\\
    \midrule
    VHS & 0.62 & 0.82 & 0.71 \\
    NACC & 0.68 & 0.59 & 0.63 \\
    ADNI & 0.86 & 0.77 & 0.81 \\
    Public only (NACC + ADNI) & 0.84 & 0.71 & 0.77 \\
    Combined & 0.72 & 0.61 & 0.66 \\
    \bottomrule
    \end{tabular}
    \label{tab:v2x-metrics}
\end{table}

\subsection{Visualization of significant brain regions}
\begin{figure*}[t]
\centering
\includegraphics[width=0.958\textwidth]{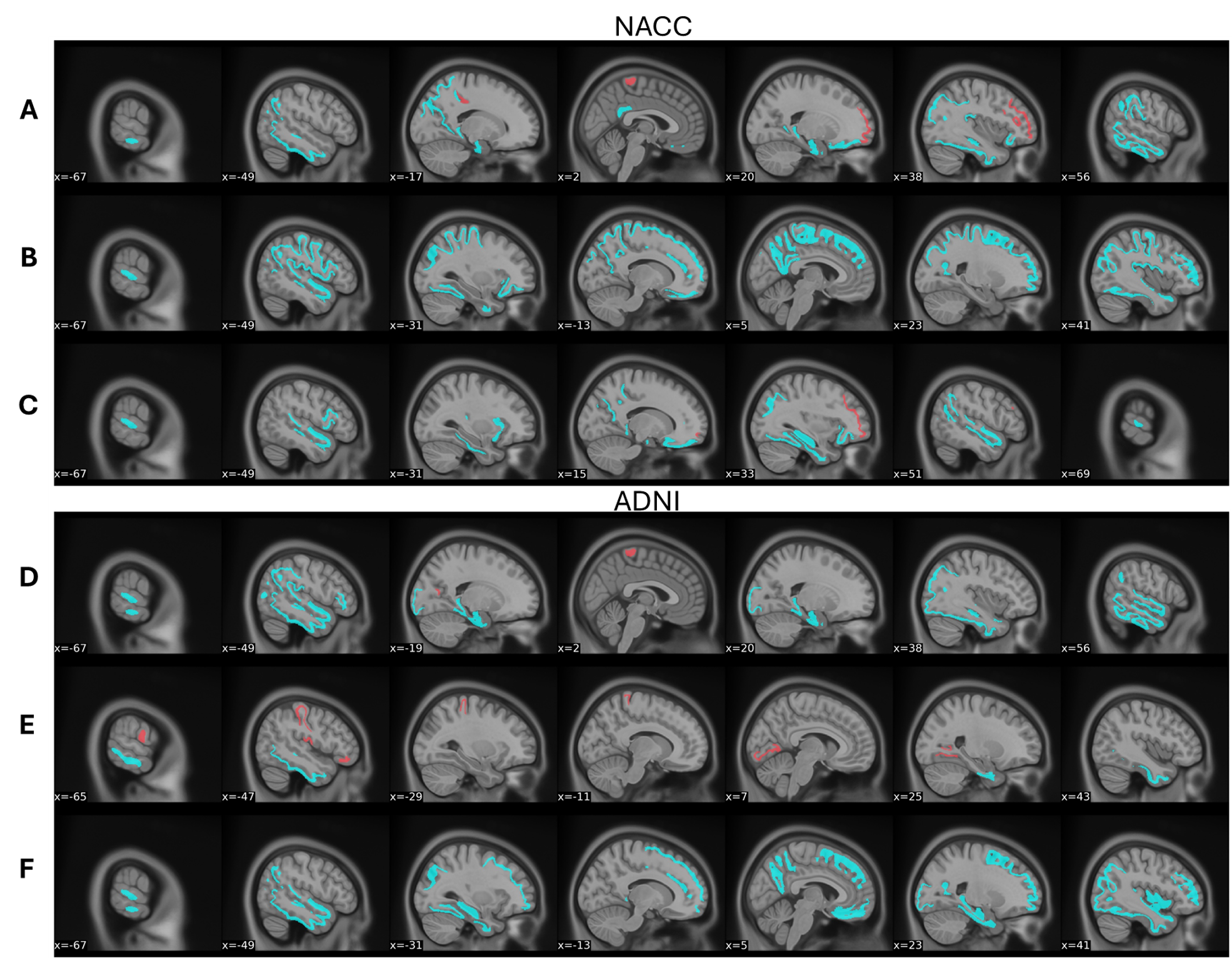}
\caption{Visualization of significant (Type I error rate $\alpha$=0.05) brain regions for \textbf{(A/D)} tAD, \textbf{(B/E)} atAD, and \textbf{(C/F)} non-AD groups compared to CN subjects within the NACC \textbf{(A-C)} and ADNI \textbf{(D-F)} data sets. Blue indicates regions of decreased volume, cortical thickness, or surface area compared to CN. Red indicates increased volume, cortical thickness, or surface area compared to CN.}
\label{fig:fig3}
\end{figure*}

Our Boruta analysis yields a selection of brain regions that are significant for distinguishing tAD, atAD, and non-AD from CN and characterizes these groups based on their patterns of brain atrophy or enhancement. Fig.~\ref{fig:fig3} visualizes the significant brain regions identified for each group within NACC and ADNI data sets. \par

For NACC, Fig.~\ref{fig:fig3}A shows that the tAD group shows significant atrophy in the hippocampal, entorhinal, fusiform, inferior parietal, inferior temporal, isthmus cingulate, lateral orbitofrontal, middle temporal, superior parietal, and supramarginal regions. The tAD group also shows increased caudal anterior cingulate volume and surface area, paracentral volume and surface area, posterior cingulate volume, and rostral middle frontal cortical thickness compared to CN. In Fig.~\ref{fig:fig3}B, the atAD group displays significant atrophy in the fusiform, inferior parietal, isthmus cingulate, lateral orbitofrontal, paracentral, pars opercularis, postcentral, precentral, precuneus, rostral middle frontal, superior frontal, superior parietal, superior temporal, and supramarginal regions. The nonAD group is characterized by significant atrophy in hippocampal, entorhinal, fusiform, inferior parietal, lateral orbitofrontal, parahippocampal, pars opercularis, precuneus, rostral anterior cingulate, superior temporal, and insular regions, as well as increased rostral middle frontal cortical thickness (Fig.~\ref{fig:fig3}C). \par

\begin{figure*}
\centering
\includegraphics[width=0.959\textwidth]{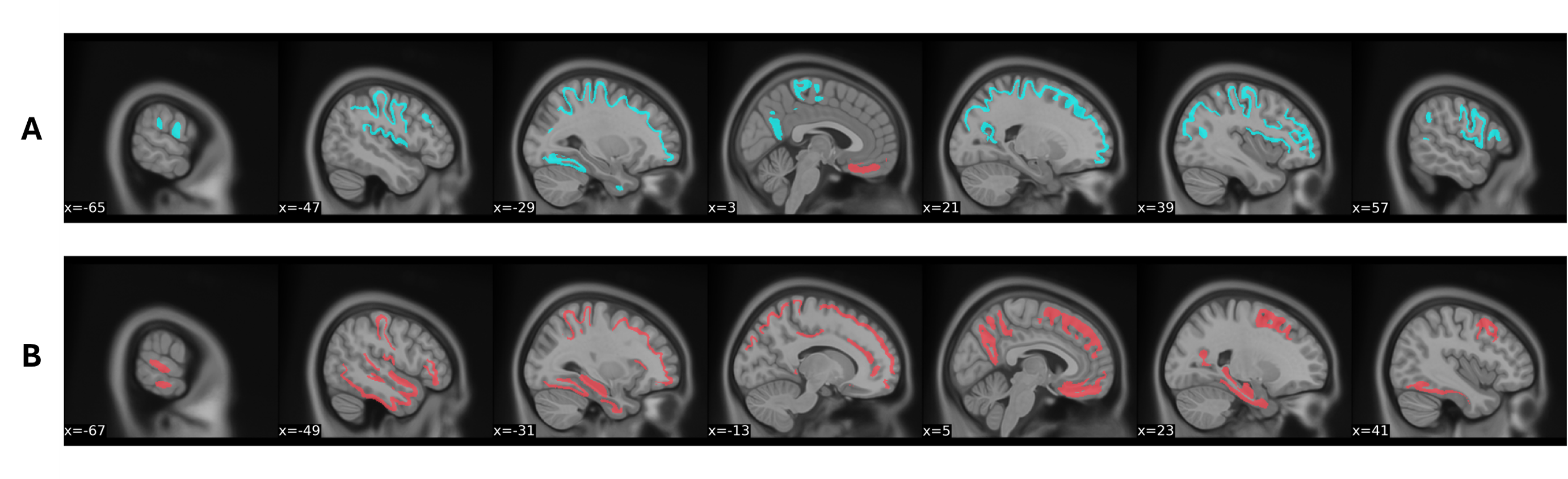}
\caption{Visualization of significant (Type I error rate $\alpha$=0.05) brain regions distinguishing between atAD and non-AD groups in the \textbf{(A)} NACC and \textbf{(B)} ADNI data sets. Blue indicates regions of decreased volume, cortical thickness, or surface area for atAD compared to non-AD. Red indicates increased volume, cortical thickness, or surface area for atAD compared to non-AD.}
\label{fig:fig4}
\end{figure*}

We perform Boruta analyses to investigate significant features distinguishing between atAD and non-AD for improved atAD diagnosis. Fig.~\ref{fig:fig4} visualizes the identified regions for atAD vs. non-AD within (A) NACC and (B) ADNI data sets. \par

For NACC, the atAD group displays significantly more atrophy than the nonAD group in the caudal middle frontal, fusiform, inferior parietal, paracentral, pars opercularis, pericalcarine, postcentral, precentral, precuneus, rostral middle frontal, superior frontal, superior parietal, and supramarginal regions (Fig.~\ref{fig:fig4}A). The nonAD group shows significantly lower medial orbitofrontal surface area compared to the atAD group (Fig.~\ref{fig:fig4}A). \par

For ADNI, the nonAD group displays significantly more atrophy than the atAD group in hippocampal, caudal middle frontal, entorhinal, fusiform, inferior temporal, medical orbitofrontal, middle temporal, parahippocampal, pars orbitalis, pars triangularis, postcentral, posterior cingulate, precuneus, rostral anterior cingulate, rostral middle frontal, superior frontal, superior parietal, and superior temporal regions (Fig.~\ref{fig:fig4}B). \par

\section{Discussion}
As demonstrated by our results, the proposed ML-enhanced approach substantially improves the diagnosis of atAD cases. For the private VHS data set, we achieve a recall of 0.82 (Table~\ref{tab:tab3}, 82\% of atAD patients correctly diagnosed) using only features currently used in clinical practice (subject demographic features, cognitive test scores, and hippocampal volume). For the public NACC and ADNI data sets, the proposed ML approach with clinical features and hippocampal volume improves diagnostic recall from 0.52 to 0.59 for NACC and from 0.34 to 0.77 for ADNI (Tables ~\ref{tab:tab2} and ~\ref{tab:tab3}). Incorporating additional MRI features, we further improve the diagnostic recall to 0.69 for NACC while achieving high precision of 0.77 (Table~\ref{tab:tab4}). For ADNI, incorporating additional MRI features with clinical features improves precision to 0.89 while maintaining a recall of 0.77 (Table~\ref{tab:tab4}) compared to clinical features and hippocampus alone. The findings suggest that inclusion of MRI features from areas of the brain in addition to the hippocampus may further improve diagnostic accuracy for atAD cases (Table~\ref{tab:tab4}). Extracting such MRI features requires no additional procedures for the patient as sagittal full brain MRI is already performed as standard of care. \par

As shown in Fig.~\ref{fig:fig3}, there is some variation in significant brain regions between the two data sets. These differences may be attributed to different stages of disease progression or disease heterogeneity (e.g., variants of atAD, different diagnoses making up the non-AD group), or the possible presence of multi-etiology dementia. The hippocampus is highlighted as a significant region for tAD (Fig.~\ref{fig:fig3}A and D) but not atAD (Fig.~\ref{fig:fig3}B and E) for both data sets, which is consistent with hippocampus-sparing nature of atAD reported in the literature. Visual observation suggests the presence of increased cortical thickness for cognitively impaired groups (tAD, atAD, and non-AD) compared to CN subjects (Fig.~\ref{fig:fig3}). Recently, Williams et al. report that cortical volume and thickness may increase in the beginning stages of disease, possibly before the development of symptoms, followed by a decrease in volume/thickness in later stages of disease \cite{RN1}. Visualization of MRI features reveals that the characteristic brain regions for non-AD group differ between NACC (Fig.~\ref{fig:fig3}C) and ADNI (Fig.~\ref{fig:fig3}F).  As shown in Fig.~\ref{fig:fig4}, variations in the atAD and non-AD groups yield data set-specific patterns of significant regions distinguishing between atAD and non-AD. Future studies incorporating larger samples of non-AD patients with non-AD diagnostic information (e.g., frontotemporal dementia) may further investigate MRI features that reliably identify atAD variants from other diagnoses contributing to cognitive impairment. Furthermore, dementia may be multi-etiology, e.g., both AD and Lewy bodies or vascular, and various combinations of pathology may result in various patterns of atrophy and compensatory strengthening effects within the brain that may further complicate diagnosis. Further comprehensive studies are needed to disentangle the effects of disease heterogeneity and co-occurring neurological conditions. \par

\section*{Data availability}
The public data sets that support the findings of this study may be requested from ADNI \cite{RN10} and NACC \cite{RN9} respectively. The VHS data are not publicly available due to restrictions of the IRB approval for this study (\#18-08-WC-0215) that are designed to ensure the privacy of the human subjects. \par

\section*{Acknowledgements}
Data collection and sharing for ADNI is funded by the National Institute on Aging (National Institutes of Health Grant U19AG024904). The grantee organization is the Northern California Institute for Research and Education. In the past, ADNI has also received funding from the National Institute of Biomedical Imaging and Bioengineering, the Canadian Institutes of Health Research, and private sector contributions through the Foundation for the National Institutes of Health (FNIH) including generous contributions from the following: AbbVie, Alzheimer’s Association; Alzheimer’s Drug Discovery Foundation; Araclon Biotech; BioClinica, Inc.; Biogen; BristolMyers Squibb Company; CereSpir, Inc.; Cogstate; Eisai Inc.; Elan Pharmaceuticals, Inc.; Eli Lilly and Company; EuroImmun; F. Hoffmann-La Roche Ltd and its affiliated company Genentech, Inc.; Fujirebio; GE Healthcare; IXICO Ltd.; Janssen Alzheimer Immunotherapy Research \& Development, LLC.; Johnson \& Johnson Pharmaceutical Research \& Development LLC.; Lumosity; Lundbeck; Merck \& Co., Inc.; Meso Scale Diagnostics, LLC.; NeuroRx Research; Neurotrack Technologies; Novartis Pharmaceuticals Corporation; Pfizer Inc.; Piramal Imaging; Servier; Takeda Pharmaceutical Company; and Transition Therapeutics. \par

The NACC database is funded by NIA/NIH Grant U24 AG072122. NACC data are contributed by the NIA-funded ADRCs: P30 AG062429 (PI James Brewer, MD, PhD), P30 AG066468 (PI Oscar Lopez, MD), P30 AG062421 (PI Bradley Hyman, MD, PhD), P30 AG066509 (PI Thomas Grabowski, MD), P30 AG066514 (PI Mary Sano, PhD), P30 AG066530 (PI Helena Chui, MD), P30 AG066507 (PI Marilyn Albert, PhD), P30 AG066444 (PI John Morris, MD), P30 AG066518 (PI Jeffrey Kaye, MD), P30 AG066512 (PI Thomas Wisniewski, MD), P30 AG066462 (PI Scott Small, MD), P30 AG072979 (PI David Wolk, MD), P30 AG072972 (PI Charles DeCarli, MD), P30 AG072976 (PI Andrew Saykin, PsyD), P30 AG072975 (PI David Bennett, MD), P30 AG072978 (PI Neil Kowall, MD), P30 AG072977 (PI Robert Vassar, PhD), P30 AG066519 (PI Frank LaFerla, PhD), P30 AG062677 (PI Ronald Petersen, MD, PhD), P30 AG079280 (PI Eric Reiman, MD), P30 AG062422 (PI Gil Rabinovici, MD), P30 AG066511 (PI Allan Levey, MD, PhD), P30 AG072946 (PI Linda Van Eldik, PhD), P30 AG062715 (PI Sanjay Asthana, MD, FRCP), P30 AG072973 (PI Russell Swerdlow, MD), P30 AG066506 (PI Todd Golde, MD, PhD), P30 AG066508 (PI Stephen Strittmatter, MD, PhD), P30 AG066515 (PI Victor Henderson, MD, MS), P30 AG072947 (PI Suzanne Craft, PhD), P30 AG072931 (PI Henry Paulson, MD, PhD), P30 AG066546 (PI Sudha Seshadri, MD), P20 AG068024 (PI Erik Roberson, MD, PhD), P20 AG068053 (PI Justin Miller, PhD), P20 AG068077 (PI Gary Rosenberg, MD), P20 AG068082 (PI Angela Jefferson, PhD), P30 AG072958 (PI Heather Whitson, MD), P30 AG072959 (PI James Leverenz, MD). \par

The NACC database is funded by NIA/NIH Grant U24 AG072122. SCAN is a multi-institutional project that was funded as a U24 grant (AG067418) by the National Institute on Aging in May 2020. Data collected by SCAN and shared by NACC are contributed by the NIA-funded ADRCs as follows: \par

Arizona Alzheimer’s Center - P30 AG072980 (PI: Eric Reiman, MD); R01 AG069453 (PI: Eric Reiman (contact), MD); P30 AG019610 (PI: Eric Reiman, MD); and the State of Arizona which provided additional funding supporting our center; Boston University - P30 AG013846 (PI Neil Kowall MD); Cleveland ADRC - P30 AG062428 (James Leverenz, MD); Cleveland Clinic, Las Vegas – P20AG068053; Columbia - P50 AG008702 (PI Scott Small MD); Duke/UNC ADRC – P30 AG072958; Emory University - P30AG066511 (PI Levey Allan, MD, PhD); Indiana University - R01 AG19771 (PI Andrew Saykin, PsyD); P30 AG10133 (PI Andrew Saykin, PsyD); P30 AG072976 (PI Andrew Saykin, PsyD); R01 AG061788 (PI Shannon Risacher, PhD); R01 AG053993 (PI Yu-Chien Wu, MD, PhD); U01 AG057195 (PI Liana Apostolova, MD); U19 AG063911 (PI Bradley Boeve, MD); and the Indiana University Department of Radiology and Imaging Sciences; Johns Hopkins - P30 AG066507 (PI Marilyn Albert, Phd.); Mayo Clinic - P50 AG016574 (PI Ronald Petersen MD PhD); Mount Sinai - P30 AG066514 (PI Mary Sano, PhD); R01 AG054110 (PI Trey Hedden, PhD); R01 AG053509 (PI Trey Hedden, PhD); New York University - P30AG066512-01S2 (PI Thomas Wisniewski, MD); R01AG056031 (PI Ricardo Osorio, MD); R01AG056531 (PIs Ricardo Osorio, MD; Girardin Jean-Louis, PhD); Northwestern University - P30 AG013854 (PI Robert Vassar PhD); R01 AG045571 (PI Emily Rogalski, PhD); R56 AG045571, (PI Emily Rogalski, PhD); R01 AG067781, (PI Emily Rogalski, PhD); U19 AG073153, (PI Emily Rogalski, PhD); R01 DC008552, (M.-Marsel Mesulam, MD); R01 AG077444, (PIs M.-Marsel Mesulam, MD, Emily Rogalski, PhD); R01 NS075075 (PI Emily Rogalski, PhD); R01 AG056258 (PI Emily Rogalski, PhD); Oregon Health \& Science University - P30 AG066518 (PI Lisa Silbert, MD, MCR); Rush University - P30 AG010161 (PI David Bennett MD); Stanford – P30AG066515; P50 AG047366 (PI Victor Henderson MD MS); University of Alabama, Birmingham – P20; University of California, Davis - P30 AG10129 (PI Charles DeCarli, MD); P30 AG072972 (PI Charles DeCarli, MD); University of California, Irvine - P50 AG016573 (PI Frank LaFerla PhD); University of California, San Diego - P30AG062429 (PI James Brewer, MD, PhD); University of California, San Francisco - P30 AG062422 (Rabinovici, Gil D., MD); University of Kansas - P30 AG035982 (Russell Swerdlow, MD); University of Kentucky - P30 AG028283-15S1 (PIs Linda Van Eldik, PhD and Brian Gold, PhD); University of Michigan ADRC - P30AG053760 (PI Henry Paulson, MD, PhD) P30AG072931 (PI Henry Paulson, MD, PhD) Cure Alzheimer's Fund 200775 - (PI Henry Paulson, MD, PhD) U19 NS120384 (PI Charles DeCarli, MD, University of Michigan Site PI Henry Paulson, MD, PhD) R01 AG068338 (MPI Bruno Giordani, PhD, Carol Persad, PhD, Yi Murphey, PhD) S10OD026738-01 (PI Douglas Noll, PhD) R01 AG058724 (PI Benjamin Hampstead, PhD) R35 AG072262 (PI Benjamin Hampstead, PhD) W81XWH2110743 (PI Benjamin Hampstead, PhD) R01 AG073235 (PI Nancy Chiaravalloti, University of Michigan Site PI Benjamin Hampstead, PhD) 1I01RX001534 (PI Benjamin Hampstead, PhD) IRX001381 (PI Benjamin Hampstead, PhD); University of New Mexico - P20 AG068077 (Gary Rosenberg, MD); University of Pennsylvania - State of PA project 2019NF4100087335 (PI David Wolk, MD); Rooney Family Research Fund (PI David Wolk, MD); R01 AG055005 (PI David Wolk, MD); University of Pittsburgh - P50 AG005133 (PI Oscar Lopez MD); University of Southern California - P50 AG005142 (PI Helena Chui MD); University of Washington - P50 AG005136 (PI Thomas Grabowski MD); University of Wisconsin - P50 AG033514 (PI Sanjay Asthana MD FRCP); Vanderbilt University – P20 AG068082; Wake Forest - P30AG072947 (PI Suzanne Craft, PhD); Washington University, St. Louis - P01 AG03991 (PI John Morris MD); P01 AG026276 (PI John Morris MD); P20 MH071616 (PI Dan Marcus); P30 AG066444 (PI John Morris MD); P30 NS098577 (PI Dan Marcus); R01 AG021910 (PI Randy Buckner); R01 AG043434 (PI Catherine Roe); R01 EB009352 (PI Dan Marcus); UL1 TR000448 (PI Brad Evanoff); U24 RR021382 (PI Bruce Rosen); Avid Radiopharmaceuticals / Eli Lilly; Yale - P50 AG047270 (PI Stephen Strittmatter MD PhD); R01AG052560 (MPI: Christopher van Dyck, MD; Richard Carson, PhD); R01AG062276 (PI: Christopher van Dyck, MD); 1Florida - P30AG066506-03 (PI Glenn Smith, PhD); P50 AG047266 (PI Todd Golde MD PhD) \par

\section*{Funding}
This research is partially supported by the Hampton Roads Biomedical Research Consortium (HRBRC) under HRBRC Grant No. 300883-002 and the National Institutes of Health (NIH) under NIH Grant No. R01EB020683. This research was supported by the Research Computing clusters at Old Dominion University, which are supported in part by National Science Foundation (NSF) Grant No. CNS-1828593. \par

\section*{Competing interests}
The authors report no competing interests. \par

\bibliographystyle{IEEEtran}
\bibliography{references}

\end{document}